# Medical Text Classification using Convolutional Neural Networks

Mark HUGHES[a], Irene LI[a,b,1], Spyros KOTOULAS[a] and Toyotaro SUZUMURA[b,c]
[a] *IBM Research Lab, Ireland*
[b] *Japan Science and Technology Agency, Tokyo, Japan*
[c] *IBM TJ Watson Research Center, New York, USA*

**Abstract.** We present an approach to automatically classify clinical text at a sentence level. We are using deep convolutional neural networks to represent complex features. We train the network on a dataset providing a broad categorization of health information. Through a detailed evaluation, we demonstrate that our method outperforms several approaches widely used in natural language processing tasks by about 15%.

## 1. Introduction

Notes are key means of recording information about the health. Health professionals spend a lot of time scanning through notes with a view on identifying key problems and getting an overall impression of the status of the person. Particularly for complex cases that lead to information overload, delays or missing information [9]. With the notable exception of the works [1], current approaches rely on dictionaries to represent meaning, but introduces limitations at the representation and modeling levels, especially when representing social determinants of health. For example, the sentence "the patient lives with their mother, who is not able to leave her home" is important in a care management setting. It is hard to model using existing tools: first, the fragment refers to the patient's mother, rather than the patient; second, it indicates social exclusion without using any word that is associated with social exclusion by itself.

The goal behind this work is to apply machine learning approaches to build models that allow an automatically generated context-based and rich representation of health-related information. Convolutional neural networks (CNNs) have dramatically improved the approaches to many active research problems. One of the key differentiators between CNNs and traditional machine learning approaches is the ability for CNNs to learn complex feature representations.

We apply a CNN-based based approach to categorization of text fragments, at a sentence level, based on the emergent semantics extracted from a corpus of medical text. We compare our approach with three other methods: Sentence Embeddings, Mean Word

---

[1] Corresponding author, Centre for Innovation, 7 Hanover Quay, Grand Canal Dock, Dublin, Ireland; Email: irenelizihui@gmail.com

Embeddings and Word Embeddings with BOW (bag-of-word). Our results indicate that the CNN-based approach is outperforming the other approaches by at least 15% in terms of accuracy in the task of classification.

## 2. Related Works

Classification for health-related text is considered a special case of text classification. Machine learning algorithms in Natural Language Processing (NLP) and have been successfully applied: e.g. Support Vector Machines and Latent Dirichlet Allocation have been used for some tasks like classification on patient record notes [1] or other documents in diseases like diabetes showing satisfying results [2,3]. The state-of-the-art models on document classification methods are designed for neural networks. Mikolov *et al*. [7] introduce an approach for learning word vector representations, Word2vec, which is simple and efficient. For neural embeddings, Le *et al*. [4] introduce the distributed representations of paragraphs, the Doc2vec, capturing the semantics in dense vectors. Other studies on CNNs for learning high level features have also shown competitive results. Other works by Kalchbrenner *et al*. [5] develop the Dynamic Convolutional Neural Networks for modelling sentences. This work is the first approach using such technology to do sentence-level classification of medical text.

## 3. Methods

We describe our method by means of a case study, where we have used Word2vec for a large corpus of text and a smaller corpus of pre-categorized text to train our sentence-level classifier.

We have used two datasets, procured from the medical domain. Our approach makes use of the Word2vec algorithm. It has been shown that performance can be improved by training Word2vec models using domain specific data. Therefore, we have procured a dataset from PubMed[1] for training our Word2vec models. To train our Word2vec models, we used a collection of 15k clinical research papers representing a wide range of medical subjects. The Word2vec model described in this paper was trained using this PubMed collection.

For sentence level classification, it was necessary to gather training data that had been pre-classified by medical professionals. Merck Manual[2] dataset contains articles from various topics like Brain, Cancer, etc. Each of these articles is classified under a parent header representing a specific category of medical issues and conditions. In total our dataset consisted of 26 medical categories and 4000 sentences were chosen at random for each category extracted from the Merck articles to use as our training data and to ensure balance across all categories. Our validation dataset consisted of 1000 sentences from each of the categories.

We apply a CNN-based approach to automatically learn and classify sentences into one of the 26 categories in our evaluation dataset. Similar to the approach outlined by

---

[1] PubMed is an online medical publication repository and contains published medical research across a very wide spectrum of clinical subjects. https://www.ncbi.nlm.nih.gov/pubmed

[2] Merck Manual is an online and offline resource containing encyclopedic style articles describing a wide range of medical subjects (http://www.merckmanuals.com/)

Kim [8], we convert each sentence to a word level matrix where each row in the matrix is a sentence vector extracted from our Word2vec model. CNNs require input to have a static size and sentence lengths can vary greatly. Therefore, we chose a max word length of 50 allowable for a sentence which worked well. During the training phase, we applied a Word2vec hidden layer size of 100, thus giving our input feature a resolution of 100×50. If a sentence contained less than 50 tokens, a special stop word was repeatedly appended to the end of the sentence to meet the 50-word requirement. If a sentence contained over 50 words, only the first 50 were considered to be representative of that sentence.

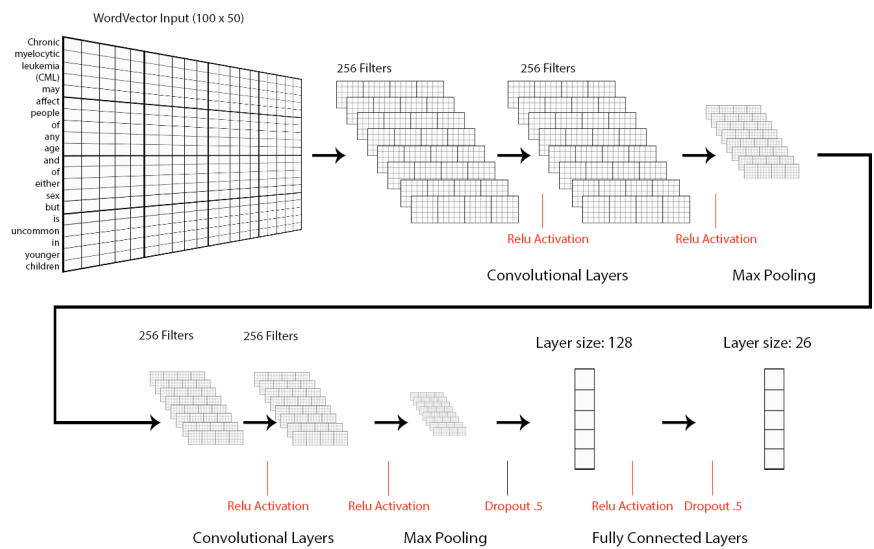

**Figure 1.** Outline of our CNN model structure.

During evaluation, we tested various CNN configurations. We applied a grid search to ascertain the optimal number of filters and filter sizes. We also experimented with multiple configurations of convolutional layers including 2, 4 and 6. From these experiments, our best performing CNN model consisted of a configuration of two sets of two convolutional layers with each pair followed by a max pooling layer. In this model, we used 256 convolutional filters with a filter size of 5 across all convolutional layers. After the second max pooling function we apply a dropout function to help preventing overfitting. In our model, we use a dropout rate of .5. We then append a fully connected layer with a length of 128 followed by a second dropout function. This is followed by a dense layer with a size of 26 to represent the number of classification classes with a Softmax function determining the output. A visual representation of this model can be found in Figure 1.

## 4. Evaluation

In this section, we evaluate our approach against a set of state-of-the-art methods. We have compared our model with the following methods: *Sentence Embeddings*, *Mean*

*Word Embeddings* and *Word Embeddings with BOW* (bag-of-word). During training word/sentence embeddings, the stop-words are kept and no stemming is adopted, since, in this way, we will keep the complete information. We keep the Doc2vec sentence embedding dimension to be 100, and the epoch to be 60 for all the experiments.

Sentence Embeddings (*LogR+Doc2vec*): Doc2vec, the *distributed memory* (PV-DM) model is firstly applied to train on the entire corpus. Once the model has been trained, each sentence in the test dataset can be inferred directly from the model. The second stage is to apply a Logistic Regression (LogR) classifier given the sentence embeddings inferred from the Doc2vec model.

Mean Word Embeddings (*ZeroMean/ElimMean+Word2vec*): For each sentence, we first took the embeddings of each word, and calculated a pair-wise mean as the sentence embedding. The lengths of sentences could be various - however, the dimension of the sentence embeddings is constant. When there is a word that is not in our vocabulary, we can fill with zeros or eliminate it (*ZeroMean+Word2vec* and *ElimMean+Word2vec* respectively). The mean word embeddings are the inputs to the above mentioned classifiers.

Word Embeddings with BOW Features (*BOW+LogR*): The third evaluation approach is the widely used Bag-of-Words histogram approach based on word vectors. As a part of this approach, Word2vec features are extracted from all entities within our dataset. We apply a k-means clustering algorithm with a value of 1000 for K to generate a feature vocabulary. Once this vocabulary is generated, a sentence is converted to a BOW histogram by assigning each word within the sentence to a vocabulary feature. Due to the short length of sentences in comparison to our vocabulary size, to avoid sparse vectors, we apply soft assignment when generating each BOW histogram. We use a value of 50 for K in the soft assignment phase with a value of 1/R appended to each histogram bin where R is the nearest neighbor ranking for the vocabulary feature associated with that bin up to a value of K.

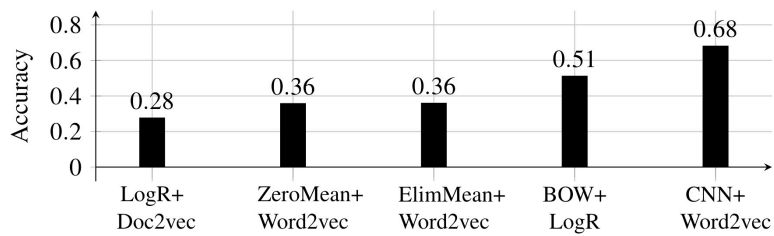

**Figure 2.** Classification Performance.

Figure 2 shows the accuracy (percentage of sentences classified correctly) of each method in our experiments. The first three methods shown in the figure give worse performance because the initially pre-trained embeddings are not providing good features for the classification. The bag-of-word method performs better - probably due to better feature extraction based on the pre-trained word embeddings. Our CNN-based approach has the highest accuracy by a wide margin. This could be explained by the fact that our deep learning approach has the ability to capture more complex features compared with the other shallow learning approaches.

## 5. Conclusions and Future Work

In this paper, we introduce a novel approach for sentence-level classification of medical documents. We show that it is possible to use CNNs to represent the semantics of clinical text enabling semantic classification at a sentence level. When compared with shallow learning methods, the multi-layer convolutional deep networks can generate more optimal features during the training phase to represent the semantics of the sentence being analyzed. Similar to computer vision methods, once these semantic representations are learned, they can also be used for many alternative tasks such as text comparison and retrieval tasks. With minimum effort, the approach could also be scaled up to generate representations at a paragraph or even document level.

In the future, we wish to implement our technique at a much larger scale and with a more fine-grained set of clinical classifications. We would expect that similar to the computer vision literature, our convolutional network approaches will provide better results when fed with larger datasets. Within the medical domain, we aim to test this hypothesis with a much larger collection of data collated from PubMed, relevant topics in Wikipedia, as well as medical books and journals. Alternatively, we also could leverage existing domain adaptation techniques [6,10] to transfer knowledge from other domains to the medical domain. We are in the process of deploying a system using a similar technique with application to Care Management. Furthermore, we aim to experiment with the feature representations generated from a patient's clinical notes and apply them to generate a high level semantic representation of each patient. A patient could then be represented as a dense and highly discriminative feature vector that captures the medical conditions and treatments from their unstructured clinical notes, possibly combined with structured data.

**Acknowledgement**
This work was partly supported by JST CREST Grant Number JPMJCR1303, Japan.